\documentclass[runningheads]{template_files/llncs}

\usepackage{template_files/eccv}

\usepackage{template_files/eccvabbrv}

\usepackage{graphicx}
\usepackage{booktabs}
\usepackage{multirow}
\usepackage{subcaption}

\usepackage[accsupp]{axessibility}  
\usepackage{amsmath}

\usepackage{hyperref}

\usepackage{orcidlink}
\newcommand\subs[1]{_{\text{#1}}}

\usepackage[normalem]{ulem}
\useunder{\uline}{\ul}{}

\makeatletter
\def\blfootnote{\xdef\@thefnmark{}\@footnotetext}
\makeatother

\begin{document}

\title{Introducing a Class-Aware Metric for\\ Monocular Depth Estimation:\\ An Automotive Perspective}

\titlerunning{Class-Aware Metric for Monocular Depth Estimation}

\author{Tim Bader\inst{1, 5}*,
Leon Eisemann\inst{2,3}*,
Adrian Pogorzelski\inst{1,4}*,
Namrata Jangid\inst{2}*,\\
and Attila-Balazs Kis\inst{2}*
}

\authorrunning{T.~Bader et al.}

\institute{Dr. Ing. h.c. F. Porsche AG, Stuttgart, Germany \\
\email{\{tim.bader, adrian.pogorzelski2\}@porsche.de}\and
Porsche Engineering Group GmbH, Weissach, Germany \\
\email{\{leon.eisemann, namrata.jangid, attila-balazs.kis\}@porsche-engineering.de}\and
Institute for Applied AI, Stuttgart Media University, Stuttgart, Germany \and
University of Freiburg, Freiburg, Germany \and
Ulm University, Ulm, Germany
}

\maketitle
\blfootnote{*These authors contributed equally.}
\begin{abstract}
The increasing accuracy reports of metric monocular depth estimation models lead to a growing interest from the automotive domain. Current model evaluations do not provide deeper insights into the models' performance, also in relation to safety-critical or unseen classes. Within this paper, we present a novel approach for the evaluation of depth estimation models. Our proposed metric leverages three components, a class-wise component, an edge and corner image feature component, and a global consistency retaining component. Classes are further weighted on their distance in the scene and on criticality for automotive applications. In the evaluation, we present the benefits of our metric through comparison to classical metrics, class-wise analytics, and the retrieval of critical situations. The results show that our metric provides deeper insights into model results while fulfilling safety-critical requirements. We release the code and weights on the following repository: \href{https://github.com/leisemann/ca_mmde}{https://github.com/leisemann/ca\_mmde}.
\end{abstract}

\section{Introduction}
\label{sec:intro}
Depth estimation is crucial for the understanding of scene geometry and provides the foundation for many downstream tasks, ranging from 3D reconstruction to navigation of robots and autonomous vehicles \cite{unidepth, depth_anything_v2}. 
Through the broad availability of camera sensors the image-based depth estimation gains traction, as an alternative to more expensive and bigger LiDAR sensors\cite{eisemann2024opendrive, eisemann_expanding_2020}.
While recent Monocular Metric Depth Estimation (MMDE) approaches have shown remarkable results, their use in automotive applications is still an ongoing research topic. Especially for the use in free space detection or trajectory planning of (highly) automated vehicles, reliable and precise distance information is needed.

However recent research often lacks interpretable or task-specific evaluation for these applications.
Currently, evaluations often focus on the overall error between ground truth and prediction, typically using metrics like Root Mean Squared Error (RMSE) and Relative Absolute Error (RelAbs)\cite{dataset_kitti_benchmark, dataset_DIODE, monodepth}.
While these allow an estimate of the overall models' performance, the metrics often fail to capture the full complexity of the task.
Further, novel MMDE models \cite{unidepth, depth_anything_v2, depthanything, patchfusion, metric3Dv1, zoedepth} incorporate a multitude of optimizations, such as the focus on universal camera inputs \cite{unidepth}, higher resolutions \cite{patchfusion}, widespread scene capabilities \cite{depthanything,metric3Dv2} and more fine-grained details \cite{marigold, depth_anything_v2}, which are not thoroughly represented as well.  
Finally through the growing popularity of large-scale models, incorporating automated labeling and diverse datasets, e.g. \cite{patchfusion}, the boundaries between in and out of distribution classes fade and the overall performance of the model can surpass the downstream performance.

To enable an interpretable metric for the evaluation of generated depth maps, with a special focus on the requirements needed in automotive applications, we present a novel multi-component metric.
In this work, we focus on the mentioned shortcomings of current evaluations by providing the following contributions: 
\begin{itemize}
    \item We introduce a novel depth evaluation metric involving class-based distance, analysis of local features, and retaining global depth consistency. 
    \item We present a comprehensive evaluation of recent state-of-the-art (SOTA) models by evaluating these on an unseen dataset.
    \item We provide an in-depth analysis of safety-critical classes derived from real-world accident data, used to weight class importance within our metric.
\end{itemize}

The paper is structured as follows: Sec.~\ref{sec:related-work} provides an overview of commonly used metrics and datasets within monocular depth estimation. Subsequently, we present our proposed metric in Sec.~\ref{sec:metric_proposal} and evaluate SOTA models in Sec.~\ref{sec:experiment}.

\section{Related Work}\label{sec:related-work}
The following section analyzes current approaches on monocular depth estimation and their respective evaluation. Further, we present commonly used datasets in these works and the accompanying benchmarks.

\subsection{Depth Estimation}

The initial advance of image classification with deep neural networks~\cite{alexnet} has quickly been adopted by works on MMDE~\cite{eigen2014depth}. 
As research in this field has advanced, metrics such as RelAbs, Relative Squared Error (RelSq), RMSE, inverse depth RMSE (iRMSE), RMSE in log space (LogRMSE), Log10 Error, Scale Invariant Log Error (SILog), Mean Absolute Error (MAE) and threshold-based Delta error $\delta < 1.25^K$ gained widespread adoption \cite{dataset_kitti_benchmark, dataset_DIODE, monodepth}.

The fact that MMDE shares similarities with other image-related tasks enables the use of backbone models, pre-trained on different computer vision tasks\cite{eigen2015predicting,depth_transfer_learning}. 
Masked image modeling~\cite{beit} and student-teacher approaches~\cite{dino_v1, dino_v2} significantly improved these backbones, which are currently used by SOTA monocular depth estimators~\cite{midas, zoedepth, depthanything, unidepth, metric3Dv2}.

Other approaches focus on the model head, e.g., AdaBins~\cite{adabins}, which utilizes Transformers~\cite{transformer} and divides different depth ranges into bins, or ZoeDepth~\cite{zoedepth}, which uses a combination of relative and metric depth and adaptively chooses a model head based on the internally classified domain. UniDepth~\cite{unidepth}'s model head contains both depth and camera modules, which, next to depth, also enables an out-of-the-box prediction of 3D points by internally estimating the parameters of the input image's camera. They report Chamfer Distance (CD) and F-score ($F_A$) \cite{metric_point_clouds} in addition to the aforementioned metrics.

Metric3D~\cite{metric3Dv1, metric3Dv2} uses canonical camera space transformations and improved learning with recurrent refinement blocks applied to the initially predicted depth. They also utilize a novel Random Proposal Normalization Loss (RPNL) instead of the Scale-Shift Invariant Loss~\cite{midas} due to its global normalization. RPNL randomly crops patches from GT and prediction, then calculates the Median Absolute Deviation Normalization~\cite{loss_median_absolute_deviation_normalization}, enhancing local contrasts.

The DepthAnything-\cite{depthanything, depth_anything_v2} framework applies similar large-scale backbone-training methods with MMDE data. Promising research also exists about diffusion based depth-estimation models with just backbones~\cite{ecodepth} and complete encoder-decoder architectures~\cite{marigold, conditioned_diff_model}.

PatchFusion~\cite{patchfusion}, also a framework, generates estimations from patches at various resolutions using both coarse and fine networks, based on existing MMDE models. These estimations are then fused together using a merging network. To cope with consistency across patches, the authors introduce a Consistency Error (CE) that calculates the Mean Absolute Error (MAE) along patches with half-resolution overlap. Additionally, they introduce a Soft Edge Error (SSE), recommended by~\cite{loss_edge, loss_edge2}, that compares the disparity difference between GT and prediction with 3x3 patches around edges. The analysis of such fine-grained details is fueled by its design to process high-resolution images, which, however, are not present in most datasets.

\subsection{Datasets}
Acquiring real-world training data for MMDE models yields a significant challenge due to hardware needs. Accurate data collection requires the use of a calibrated LiDAR, camera system or stereo camera setups. 
Further, LiDAR point clouds are often sparse, while stereo metric depth is often limited in range and precision. 
Additionally, generalization through a dataset is hard due to the ill-posed nature of models having to work with unknown camera intrinsics~\cite{adabins, metric3Dv2}.

Existing datasets and benchmarks overcome those issues with sophisticated post-processing steps and by promoting common camera parameters. Prominent examples of MMDE benchmark datasets are NYU Depth~\cite{dataset_NYU_Depth_V2} and KITTI~\cite{dataset_KITTI, dataset_kitti_benchmark}, which focus on metrics such as SILog, iRMSE, RelAbs and RelSq.
Alongside those, more recent popular datasets exist ~\cite{dataset_DIODE, dataset_BDD100K, dataset_Cityscapes, dataset_nuscenes, dataset_Waymo, dataset_DDAD, dataset_goose}, containing camera data in HD to Full-HD. 
For higher resolution needs, the synthetic UnrealStereo4K~\cite{dataset_UnrealStereo4K} dataset and the Middlebury~\cite{dataset_middlebury} benchmark are available.

Synthetic datasets are another way to avoid hardware drawbacks. However, they contain a distribution shift to real-world data often leading to generalization issues. 
Notably, DepthAnything V2 \cite{depth_anything_v2} overcomes this shift by increasing the backbone size by switching from ViT-Large to ViT-Giant and thus improving its level of detail and accuracy. 
Because of the increased depth sharpness, the authors focus on the Gradient Matching Error (GME)~\cite{midas}. Alongside, they provide the novel DA-2K benchmark, but still report SOTA metrics RelAbs, RMSE, Log10, and Delta $K\in\{1,2,3\}$ errors.

\section{Metric Proposal}\label{sec:metric_proposal}
In the following section, we introduce our proposed depth estimation metric. To achieve a comprehensive evaluation of diverse scenes, our metric compromises three different levels of granularity. First, we make use of an object classification-based component, to thoroughly gather information about the models' performance over diverse, possible out-of-distribution classes. Second, we assess the models' performance to distinguish object features by leveraging e.g., edge or corner detection filters. Finally, to enable global consistency we further incorporate standard depth estimation evaluation methods. 

Within the individual components, we decided on MAE as a foundation.
\begin{equation}
\text{MAE} = \sum_{i=1}^{D}|x_i-y_i|
\end{equation}
We selected this approach because it captures average model performance errors without bias towards outliers, enabling a robust, symmetrical, and interpretable metric for across-the-board evaluations.

\subsection{Class-Based Component}\label{subsec:class-based_metric}
As described in Sec.~\ref{sec:related-work}, recent models are trained and evaluated on a wide variety of datasets focused on different use cases. Therefore, the predicted depth maps over different models can react differently to previously rare or unseen classes. 
Based on this, we introduce a class-based error measurement. Within this measurement, we evaluate the metric error of each object class, e.g. car, truck, building, pole individually.

\subsubsection{Intra-Class Weighting}\label{subsubsec:intra-class_weight}
However, we note that the importance of a class can vary highly between frames and situations. Since we focus on classification masks and not instance masks, one mask may span over a multitude of car instances both close and far in the scene. Weighting these similar to one vehicle close to the camera would bring difficulties in the interpretation of the metric. Therefore weighting the classes is necessary. Further, this weighting can not be based on the pixel area of the class in the frame, as these could lead to the same weighting in the provided example.
Consequently, we propose a distance-based intra-class weighting $w\subs{dist}$, based on the distances within each scene. We define this as
\begin{equation}
w\subs{dist} = \frac{d\subs{class} - \min(D\subs{classes})}{\max(D\subs{classes}) - \min(D\subs{classes})}
\end{equation}
\begin{equation*}
\text{with   } d\subs{class} = d\subs{scene-max} - d\subs{class-min}
\end{equation*}
where $\text{d}\subs{scene-max}$ describes the maximum distance within the entire scene and $\text{d}\subs{class-min}$ the minimum distance within a class. 
Both distances are derived from the ground truth data of the scene to prevent a model with a trained maximum distance from influencing the weighting. $D\subs{classes}$ describes the set of all $d\subs{class}$ in each image.
This simplistic approach weights classes close to the camera higher than far away objects, while incorporating the overall scenery and allowing a unified method for diverse depth imagery.

\subsubsection{Inter-Class Weighting}
Additionally to scaling the class importance in relation to the scene, not all classes have the same relevance between different use cases. To achieve a unified score for the class accuracy of the depth prediction, we introduce $w\subs{class}$ an inter-class weighting.

Since the class importance heavily relies on the use case at hand, the specific weighting of the classes can be chosen individually. As our focus is the use of MMDE models in automotive applications respectively \textit{automotive safety}, we provide an in-depth weight setup in respect thereof.

To the best of our knowledge, there is no broadly accepted class-wise importance for object detection in the automotive area. Therefore, we leverage accident data and use the distribution between the accident opponent. We source our data from the German In-Depth Accident Study (GIDAS) \cite{GIDAS} database.
\mbox{GIDAS} represents a continuing research effort aimed at enhancing road safety through the meticulous collection and analysis of traffic accident data starting from 1999. GIDAS maintains an extensive repository of data encompassing numerous parameters such as accident dynamics, vehicle and infrastructure conditions, and injury patterns. 
In our analysis, we implemented several filters to focus on the most relevant cases. 
We analyzed fully reconstructed accident data collected up until December 2022. Only accidents involving at least one injured occupant and/or an injured vulnerable road user (VRU) were included. 
Furthermore, we concentrated on post-NCAP ego vehicles, analyzing exclusively the first collision in each accident.
\begin{table*}[t]
\centering
\begin{tabular}{@{}ll@{}cc@{}}
\toprule
\textbf{Main Class}                     & \textbf{Sub Class}    & \textbf{Distribution} \\ \midrule
Car-to-Vehicle                   &          & \textbf{62,06 \% }            \\
                                 & Car         & 50,04 \% \\
                                 & Motorcycle         & 7,38 \% \\
                         & Truck   \& Van \& Bus  & 3,73 \%             \\
                         & Trains & 0,63 \% \\
                         & Other Motorized Vehicle & 0,27 \% \\
                         
Car-To-VRU &          & \textbf{30 \%}             \\
&Bicycles &  21,95\% \\  
&Pedestrian                        & 8,05 \%              \\
Car-To-Object &          & \textbf{7,94 \%}             \\
&Pole/tree                        & 3,24 \%              \\
&Guardrail                        & 1,17 \%              \\
&Ditch/   Embankment              & 1,07 \%              \\
&Road/   Terrain                  & 1,04 \%              \\
&Other   Object                   & 0,75 \%              \\
&Wall/   bridge                   & 0,56 \%              \\
&Bush/Fence                       & 0,11 \%              \\
 \bottomrule
\end{tabular}
\caption{GIDAS Distribution of accident opponents used to weight the class importance for the final metric result.}
\label{tab:dist-accid-coll}
\end{table*}
Through this, we identified a total of 22385 accidents. 
The distribution of these accidents is as follows: 
62,06 \% involved car-to-vehicle collisions, 30 \% involved VRUs and 7,94 \% involved car-to-object collisions. A detailed breakdown of these statistics appears in Tab.~\ref{tab:dist-accid-coll}. 
We make direct use of this statistical evaluation by defining our class weights $w\subs{class}$ as the presented percentages per class.

\subsubsection{Component Result}
The final class-based component is calculated using MAE, the intra-class weight $w\subs{dist}$, and the inter-class weight $w\subs{class}$.
\begin{equation}
\text{E}_{\text{class}} = \sum_{c=1}^{C} w\subs{class} \cdot w\subs{dist} \cdot \text{MAE}(I)
\end{equation}
Achieving an error $E\subs{class}$ that incorporates how important a class is in general and also how relevant this class is in the respective image situation. One should note the difference between the theoretical formula and the implementation, where the safety classes are mainly considered so-called super-classes, incorporating specific dataset classes. E.g., other motorized vehicle super-class can contain the classes: heavy machinery and kick scooters. In such cases, the sum of intra-class weighted errors for the two classes will be multiplied by the specified inter-class weight value.

\subsection{Local Feature Component}
Another important factor for a qualitative depth map is preserving fine details in the prediction. These details serve multiple purposes, such as better differentiation between individual objects or considering unique - and often relevant - shape changes such as trailer hitches or opened doors on cars.

\subsubsection{Feature Extraction}
For the task of extracting possibly relevant features, we apply several classical methods on the unmasked input image, resulting in feature map $F$. We implement a set of feature extractors, each one providing different maps, with different focuses. On one hand, a main edge detection algorithm allows the extraction of detailed object contours for contiguous areas such as vehicle windows and road markings \cite{FE_BorderFollowing}. On the other hand, we implement multiple corner detection algorithms, e.g., Harris, given the proven robustness of corner features for computer vision tasks such as feature matching. Since both methods result in strictly the feature pixels on the applied image, we provide a parameter to extend the area of interest. In the case of edge detection this parameter can be understood as border thickness around the feature pixels. In case of corner detection, as the radius of the circle with the feature pixel as center point.

To further evaluate class-specific differences in the models in question we mask the edge depth map with the previously defined classes, similar to Sec.~\ref{subsec:class-based_metric}.

\subsubsection{Component Result}
Also, the importance of edge features is dependent on the distance to the capture point, these are scaled by the $w\subs{dist}$ as described in Sec.~\ref{subsubsec:intra-class_weight}. We calculate the final feature component through
\begin{equation}
\text{E}_{\text{feature}} = \sum_{c=1}^{C} w\subs{class} \cdot w\subs{dist} \cdot \text{MAE}(I\subs{cf})
\end{equation}
where $I\subs{cf}$ describes the edge features $F$ within a mask of class $C$. It is important to note that $F$ is calculated on the unmasked input in the previous section, to preserve image gradients calculated in the process.

\subsection{Global Consistency Component}
As we aim for a comprehensive evaluation we further examine the global consistency of the generated depth map. In addition, this also covers situations in which no labels or masks for certain objects are provided, as well as global scaling issues not represented in the other components. Therefore we simply calculate $\text{E}\subs{global}$ the MAE between the predicted and ground truth depth.

\subsection{Overall Metric Conclusion}
Considering that our metric consists of multiple components focused on different characteristics of depth maps and their generation, we also report each component individually. While this provides an exhaustive insight into the quality of the depth map at hand, for direct comparison of MMDE models a single value is more advantageous. Although the individual weighting can be dependent on the specific scenario, we propose the overall combination of components as 
\begin{equation}
    L = \gamma \cdot \text{E}\subs{class} + \gamma \cdot \text{E}\subs{feature} + \gamma \cdot \text{E}\subs{global}
    \label{eg:overall_metric_result}
\end{equation}
with $\gamma = 1 $ allowing a near metric offset evaluation while incorporating the class and distance weightings. In the case of evaluation over full datasets, first, the individual components are calculated as the MAE overall image and ground truth pairs. Second, the sum of the components is combined according to Eg.~\ref{eg:overall_metric_result}.

\section{Experimental Setup}\label{sec:experiment}
 
In this section, we first explain how we source the dataset collection and analyze it. Second, we describe the GOOSE~\cite{dataset_goose} dataset which we use for our evaluation. Finally, we outline the procedure of our evaluation, in which we consider the models AdaBins~\cite{adabins}, DepthAnything V2~\cite{depthanything}, EcoDepth~\cite{ecodepth}, Marigold~\cite{marigold}, Metric3D V2\cite{metric3Dv2}, PatchFusion~\cite{patchfusion}, UniDepth V[1-2]~\cite{unidepth} and ZoeDepth~\cite{zoedepth}.

\subsection{Dataset Analysis}
Out of the different training datasets used by the evaluation models, we identify 36 depth-related ones in total \cite{dataset_3D_Movies, dataset_A2D2,dataset_ApolloScape,dataset_Argoverse2,dataset_BDD100K,dataset_BlendedMVS,dataset_Cityscapes,dataset_DDAD,dataset_DIML,dataset_DIML-Indoor,dataset_DIODE,dataset_Driving_Stereo,dataset_DSEC,dataset_HM3d,dataset_HRWSI,dataset_Hypersim,dataset_IRS,dataset_KITTI,dataset_Lyft,dataset_Mapillary_PSD,dataset_Matterport3d,dataset_MegaDepth,dataset_MVS-Synth,dataset_NYU_Depth_V2,dataset_Pandaset,dataset_ReDWeb,dataset_Replica,dataset_ScanNet,dataset_TartanAir,dataset_Taskonomy,dataset_UASOL,dataset_UnrealStereo4K,dataset_Virtual_KITTI,dataset_Virtual_KITTI_2,dataset_Waymo,dataset_WSVD}. In the context of this work, we consider datasets used for backbone training or student-teacher approaches out of scope.

\begin{figure}[tb]
  \centering
  \includegraphics[width=\linewidth]{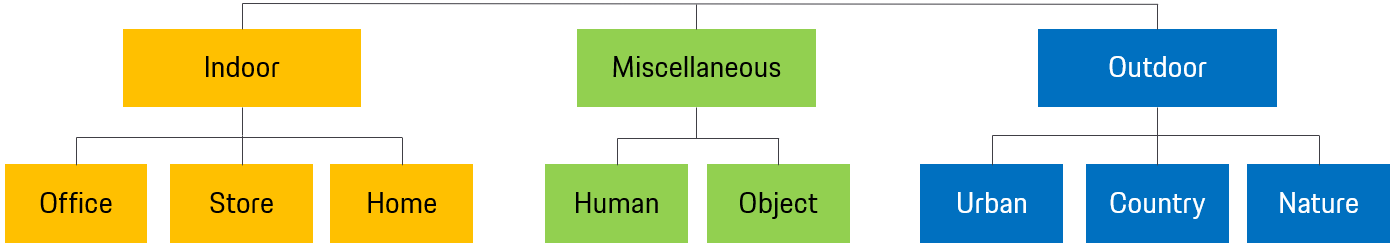}
  \caption{Class hierarchy to categorize datasets, consisting of high- and mid-level classes.}
  \label{fig:class_hierarchy}
\end{figure}

We review each dataset's technical report to collect information about the data acquisition processes, total frame count, whether the data was primarily indoor or outdoor, and the main high-level scenario classes. The estimated frame counts are then cross-verified with dataset and model reports.
For our comparative analysis, we classify the datasets into a hierarchical structure shown in Figure~\ref{fig:class_hierarchy}.
The classes \textit{Human} and \textit{Object} are human- and object-focused and allow different types of background. The \textit{Urban} class can contain lots of humans too, but in the form of pedestrians and not as the primary focus. Other classes are self-explanatory. We also take possible overlaps into account.

Because some datasets contain multiple classes, we distribute the frame count equally among the relevant classes. Data is aggregated by grouping the frame counts according to these classes. Let $C$ be the set of classes and $D_c$ the set of datasets that include class $c$. The frame count for each class is calculated as:

\begin{equation} 
\label{eq:classgrouping} 
N_{c} = \sum_{i\in C_c}\frac{|F_{i}|}{|C_{i}|} 
\end{equation} 

where $N_{c}$ represents the total frames for each class $c$, $|F_{i}|$ is the frame count of dataset $i$ and $|C_{i}|$ denotes the number of classes attributed to dataset $i$. 
With this, we calculate the share $p_{c}$ of every class in percentage:

\begin{equation}
    \label{eq:class_weighting}
    p_c = 
    \frac
        {N_{c}}
        {\sum_{c' \in C} N_{c'}} 
\end{equation}

\begin{figure}[tb]
  \centering
  \includegraphics[width=\linewidth]{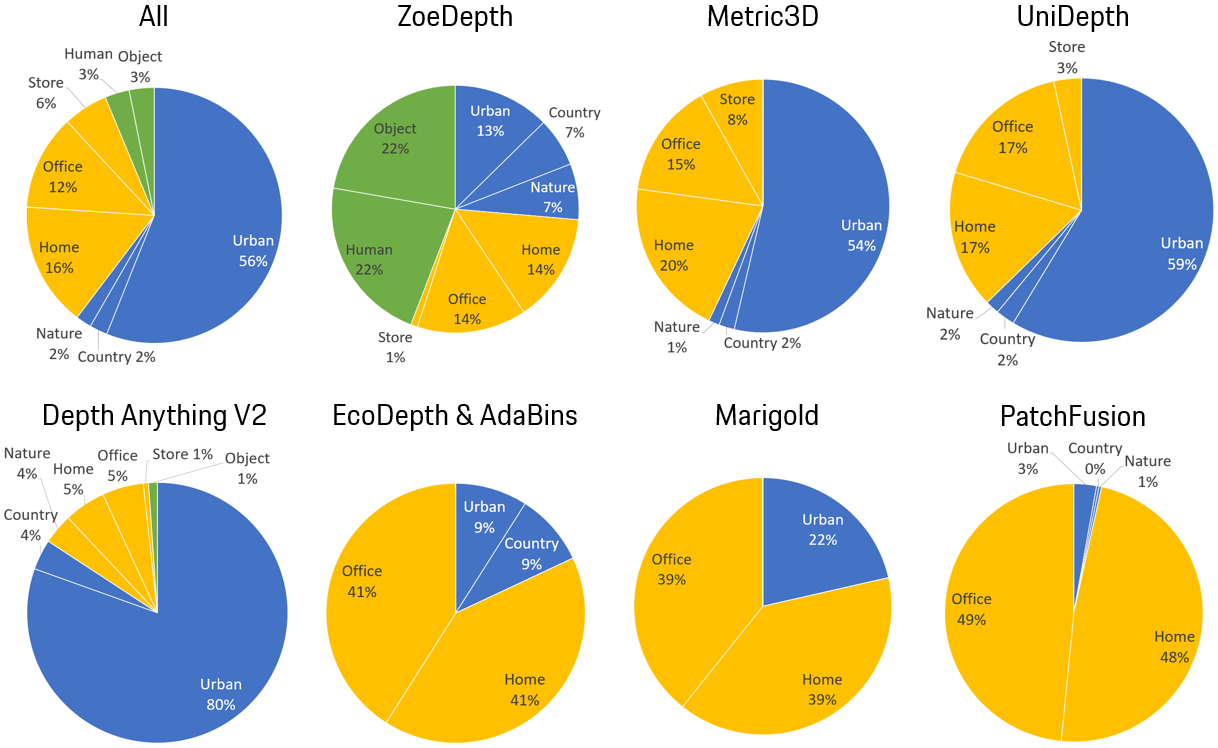}
  \caption{Distribution of the pre-defined classes that are present in the datasets used by the evaluation models. The percentages were calculated with Equation~\ref{eq:class_weighting}.}
  \label{fig:class_evaluation_all_models}
\end{figure}

We apply this over the complete dataset collection and for datasets used by models respectively and provide the results in Figure~\ref{fig:class_evaluation_all_models}. 
The acquired data shows a significant difference in the data class distribution across the models. While PatchFusion has about 97 \% of training data from indoor scenarios due to the NYU Depth V2 pre-training, DepthAnything V2 reaches just 11 \%. ZoeDepth, depending on its exact type, has the most balanced data distribution between all three high-level classes. Metric3D is evenly distributed between outdoor and indoor as well, but lacks closeup data from humans and other objects, similar to the majority of models. 

In contrast, EcoDepth and AdaBins just use KITTI and NYU Depth V2 for training, leading to the lowest amount of training data in comparison. Only a few models like ZoeDepth, EcoDepth, and AdaBins capture nature and country-like data to a significant degree. However, some of the Urban datasets contain nature parts too, depending on the respective cities.

\subsection{GOOSE Dataset}
The GOOSE \cite{dataset_goose} (German Outdoor and Offroad) dataset was designed to enhance the development and evaluation of deep learning models in unstructured outdoor environments. It offers a comprehensive collection of pixel-wise annotations of RGB images and LiDAR point clouds with 64 object classes, which enables the targeting of many out-of-distribution classes for our evaluation.

We prepare the data by extracting image-segmentation pairs from the windshield camera of the MuCAR-3 provided in \textit{ROS1} \cite{ros_stanford} bags. Notably, the segmentations are sparse and not present in every sequence. Concurrently, we extract the corresponding LiDAR point clouds that have the closest recording timestamps to the image-segmentation pairs. Since the point clouds have a frame rate of approximately 10 FPS and are provided asynchronously to the camera frames, we further extract the GPS position for each image-segmentation pair and point cloud. As GPS data is recorded at 100Hz, we compensate the positional deviations caused by time discrepancies. This is achieved by applying the translation vectors between the GPS positions to the point clouds. Afterward, we use the provided projection matrix to generate depth maps. Because the depth maps are sparse, additional interpolation is applied and the sky is masked out.

The GOOSE dataset currently does not offer a training and validation split. Therefore we randomly choose 25 scenes with a total of 1080 images. The images are provided in RGB format with a resolution of $1000\times2048$ pixels. The depth maps have source resolution and contain per-pixel distance values in meters.

\subsection{Model Evaluation Setup}\label{subsec:model-eval-setup}
To evaluate the models under similar conditions, we do not modify the input image resolution and instead interpolate the resulting depth maps to the original size when required. 
Most models offer different backbone sizes. Hence, we focus on the best-performing variant, as long as the weights are publicly available.

Models, such as UniDepth V[1-2] and Metric3D, that can process camera intrinsics, are provided with it respectively. PatchFusion offers configuration of the input image resolution and the tiling strategy, for which we use the parameters from their 2K resolution example provided on their GitHub repository.
For DepthAnything V2 we choose the Virtual KITTI checkpoint, and for AdaBins and EcoDepth the KITTI checkpoint. EcoDepth additionally requires the Stable Diffusion v1-5 pruned EMA-only encoder.

Because Marigold produces affine-invariant depth, we determine the scale and shift differences to the metric system. We do this by regressing the  function
\begin{equation}
    \label{eq:affine_invariant_regression}
    y = \text{scale} \times x + \text{shift}
\end{equation}
with $x$ being affine-invariant depth and $y$ being the correct metric depth. We use the $x$ and $y$ values of the first frame from sequence \textit{0\_Asphalt\_and\_Gravel\_Path along\_Grassland}. We then scale and shift all Marigold predictions accordingly. 

\section{Results}

\begin{table}[t]
\centering
\begin{tabular}{@{}cccccc@{}}
\toprule
{\textbf{Model}} & {\textbf{Variant}} & { \textbf{MAE}} & { \textbf{RMSE}} & {\textbf{Abs-Rel}} & {\textbf{Ours}} \\ \midrule
AdaBins              & KITTI                  & 13,3               & 25,21               & 0,33                   & 20,65               \\
DepthAnything V2    & ViT L                  & 8,39               &{\ul 16,56}               & 0,3                    & 14,47               \\
EcoDepth             & -                      & 10,25              & 20,51               & 0,28                   & 17,43               \\
Marigold             & -                      & 12,70              & 20,38               & 0,65                  & 17,72               \\
Metric3D V2         & ViT G2                 & \textbf{6,47}      & \textbf{14,44}      & \textbf{0,2}           & \textbf{11,57}      \\
PatchFusion           & DA V1 ViT L            & 15,05              & 24,33               & 0,55                   & 23,32               \\
UniDepth V1          & ConvNext L             & {\ul 8,26}               & 16,7                & {\ul 0,24}       & {\ul 14,19}               \\
UniDepth V2          & ViT L                  & 8,57               & 20                  & 0,27                   & 14,24               \\
ZoeDepth             & NYU + KITTI            & 9,51               & 19,32               & 0,27                   & 16,22               \\ \bottomrule
\end{tabular}
\caption{Comparison of the results over 25 GOOSE dataset scenes with classical errors against our metric. While both provide comprehensive insights into the model performances, ours offers a more nuanced interpretation.}
\label{tab:example_result_table}
\end{table}

In the following section, we compare our metrics results against common ones, to evaluate the benefits of our approach. From the variety of metrics presented in Sec.~\ref{sec:related-work}, we decided on MAE, RMSE, and Abs-Rel errors, because of their high spread throughout the works. For the remainder of this work, we refer to these as classical metrics. We evaluate the models defined in Sec.~\ref{subsec:model-eval-setup} on the previously selected 25 scenes. The accumulated results are presented in Tab.~\ref{tab:example_result_table}.

In the direct comparison between the individual models, Metric3D V2\cite{metric3Dv2} using the ViT G2 backbone, shows the overall best accuracy on the classical metrics. Accordingly our metric shows consistent results, indicating the same. On this, we conclude, that our metric does not negatively influence the overall performance rating of the evaluated methodologies.
Contrary the classical metrics are not aligned when focusing on the second best performing model. Here MAE, Abs-Rel, and our metric attest UniDepth V1 \cite{unidepth} the best performance, while DepthAnything V2 \cite{depth_anything_v2} achieves the lowest RMSE score. This difference highlights the interpretability of our approach. Considering the global working principle of the classical metrics and finding an exact reason for the mismatch between the metrics is challenging. In contrast, our metric allows a deeper examination of the classes and factors leading to the result, which we present in the subsequent sections.

\subsection{Single Class Evaluation}
As the key mechanics of our metric is class-based quantification, we investigate the model performance on a single class. We assume that, although models are trained on the entire images, they still can show signs of class distribution shifts.

We extract the class-based component results for the \textit{traffic signals} super-class. Traffic lights and signs were selected because they belong to the sub-class pole/tree, see Sec.~\ref{subsec:class-based_metric}, which are relevant for vehicle safety, and further crucial for autonomous driving. In other words, the class weighting considers the set of traffic signs and traffic lights as 100\% in the class-based component. The respective results are presented in Tab.~\ref{tab:example_result_table_class_eval}. According to the overall evaluation, Metric3D V2 with ViT G2 achieves the lowest error in all metrics. However, comparing the second-lowest scores shows a nonspecific result.

Within highly automated driving functions traffic signs close to the ego vehicle have a higher probability to influence the respective system decisions. In this context, our proposed intra-class weighting suggests that for closer range traffic signs UniDepth V2 \cite{unidepth} is better suited than others. 
To evaluate this assertion in practice, we investigate out-of-domain or outlier images within the dataset.

\begin{table}[t]
\centering
\begin{tabular}{@{}cccccc@{}}
\toprule
{\textbf{Model}} & {\textbf{Variant}} & { \textbf{MAE}} & { \textbf{RMSE}} & {\textbf{Abs-Rel}} & {\textbf{Ours}} \\ \midrule
AdaBins              & KITTI                  & 14,87               & 29,7               & 0,34                   & 52,55               \\
DepthAnything V2    & ViT L                  & 11,16               & 22,32               & 0,36                    & 42,59               \\
EcoDepth             & -                      & 13,28              & 26,61               & 0,32                   & 50,55               \\
Marigold             & -                      & 14,49              & 22,00               & 0,72                  & 39,53               \\
Metric3D V2          & ViT G2                 & \textbf{7,93}      & \textbf{18,74}      & \textbf{0,23}           & \textbf{37,58}      \\
PatchFusion           & DA V1 ViT L            & 17,1              & 29,03               & 0,6                   & 56,93               \\
UniDepth V1          & ConvNext L             & 10,92               & {\ul 21,5}                & {\ul 0,28}             & 38,95               \\
UniDepth V2          & ViT L                  & {\ul 10,25}              & 26,92                  & 0,33                   & {\ul 38,77}               \\
ZoeDepth             & NYU + KITTI            & 11,76               & 24,9               & 0,31                   & 44,41               \\ \bottomrule
\end{tabular}
\caption{Method results, considering only the super-class \textit{traffic signals}, comprised by the classes: traffic sign and traffic light.}
\label{tab:example_result_table_class_eval}
\end{table}

\subsection{Qualitative Metric Evaluation}
Retrieving challenging scenes from a large dataset is a complex and demanding task. As our metric incorporates multiple safety-critical aspects, we demonstrate the identification of complex scenes. Similar to previous experiments, we examine our metrics results against classical metrics, such as MAE.

\begin{figure}[t]
  \centering
  \begin{subfigure}[c]{0.32\linewidth}
    \includegraphics[width=\linewidth]{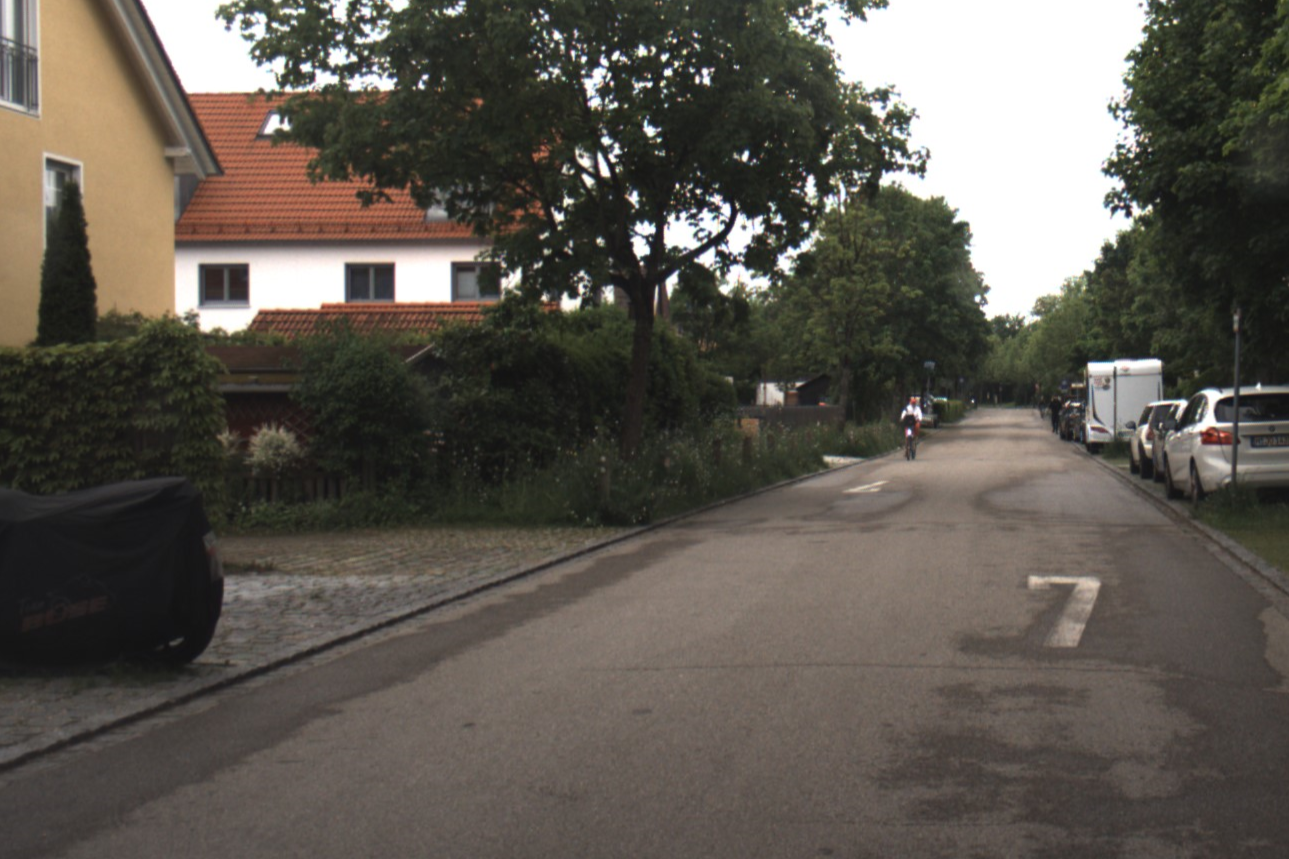}
  \end{subfigure}
  \hfill
  \begin{subfigure}[c]{0.215\linewidth}
    \includegraphics[width=\linewidth]{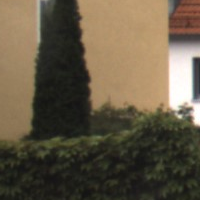}
  \end{subfigure}
  \begin{subfigure}[c]{0.215\linewidth}
    \includegraphics[width=\linewidth]{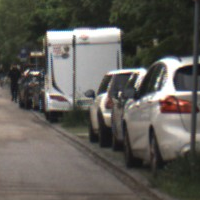}
  \end{subfigure}
  \begin{subfigure}[c]{0.215\linewidth}
    \includegraphics[width=\linewidth]{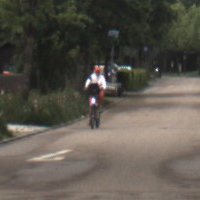}
  \end{subfigure}
  \begin{subfigure}[c]{0.32\linewidth}
    Metric 3D V2 VG2
  \end{subfigure}
  \hfill
  \begin{subfigure}[c]{0.215\linewidth}
    \includegraphics[width=\linewidth]{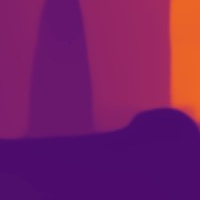}
  \end{subfigure}
  \begin{subfigure}[c]{0.215\linewidth}
    \includegraphics[width=\linewidth]{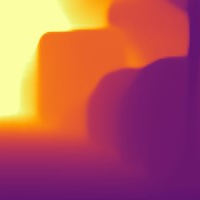}
  \end{subfigure}
  \begin{subfigure}[c]{0.215\linewidth}
    \includegraphics[width=\linewidth]{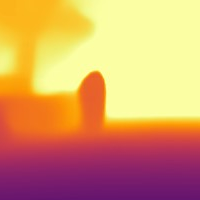}
  \end{subfigure}
  \begin{subfigure}[c]{0.32\linewidth}
  DepthAnything V2
  \end{subfigure}
  \hfill
  \begin{subfigure}[c]{0.215\linewidth}
    \includegraphics[width=\linewidth]{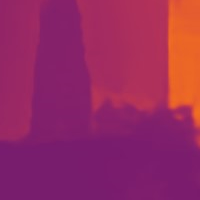}
    \caption{}\label{fig:qual-example-col-a}
  \end{subfigure}
  \begin{subfigure}[c]{0.215\linewidth}
    \includegraphics[width=\linewidth]{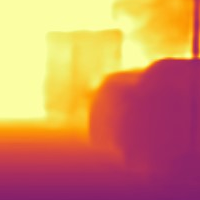}
    \caption{}\label{fig:qual-example-col-b}
  \end{subfigure}
  \begin{subfigure}[c]{0.215\linewidth}
    \includegraphics[width=\linewidth]{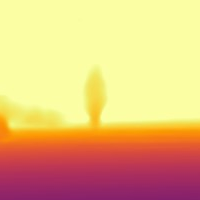}
    \caption{}\label{fig:qual-example-col-c}
  \end{subfigure}
  \caption{Example use of our metric in identifying challenging scenes for depth estimation. A classical MAE evaluation shows 3.77 for Metric3D V2 and 6.00 for DepthAnything V2, missing factors needed in safety-critical use. In comparison our Metric yields 29.97 for Metric3D V2 and 28.98 for DepthAnything. Our metric hereby weights in missed objects (a), object distinction (b), and shape representation (a) \& (c).}\label{fig:qual-example-comparison}
\end{figure}

One exemplary excerpt of our findings is displayed in Fig.~\ref{fig:qual-example-comparison}. Based on the MAE results of 3.77 for Metric3D V2 and 6.00 for DepthAnything V2, one would assume the consistently high-performance Metric3D shows throughout this work. However, our metric yields a score of 29.97 for Metric3D V2 and 28.98 for DepthAnything V2 and therefore shows contrary performance implications.

The provided cropouts of the predicted depth maps support our metrics result and highlight our safety-critical evaluation. While Metric3D V2 achieves a stable distance estimation, the representation of objects is falling short. In direct comparison, Metric3D V2 does not distinct highly occluded objects correctly from the background as shown in Fig.~\ref{fig:qual-example-col-a}. Similar phenomena are shown in Fig.~\ref{fig:qual-example-col-b}. While DepthAnything V2 can provide clear contours on the pole and the car mirror in front of the camper van, Metric3D V2 is unable to do so. In the context of automated driving functions, these details are highly relevant as the respective functions must incorporate them for obstacle detection, trajectory planning, and pre-crash estimations. Fig.~\ref{fig:qual-example-col-c} further shows better shape representation in the DepthAnything V2 prediction, as the bicyclist and the nearby grass are correctly detailed. Metric3D V2 thereby cannot distinguish the grass and tree section and predicts a wall-like structure.

The cases confirm our proposal’s working principle, as our metric accurately incorporates missed objects, object distinction, and shape representation, allowing for more reliable model weighting in safety-critical applications.

\section{Conclusion}
The growing capabilities of metrical monocular depth estimation methods have increased the importance of such applications within highly automated vehicles. However, current evaluations of these approaches fall short of the granular requirements within these safety-critical applications, as class accuracies or out-of-domain classes are not fully reflected. To incorporate this information in the review of model performances, we proposed a new metric. In contrast to classical methods such as MAE, RMSE, or Abs.-Rel., we examine on a per-class level and feature level, while preserving global consistency. Within the class component, classes are first weighted based on their distance toward the camera's principal point and second on their overall relevance for critical driving situations. While the inter-class weights can be user-defined in dependence on the use case, we provide extensive weights based on traffic accident data extracted from the GIDAS database. As detailed estimations matter, such as clear object borders or small parts such as car mirrors or hinges, local image features on corners and edges are extracted, related to the distance values, and weighted accordingly for each class. To acknowledge potential missing classes, an additional global MAE error is weighted in the final metric result, further preserving global consistency. In the evaluation, we leverage the GOOSE dataset as a granular annotated source, not included in the training of current SOTA Depth Estimation models.

Through the evaluation of a vast number of SOTA models, we provided evidence of the benefits of our proposed metric. While we could show the consistency in general model evaluation, we further showed the class-wise investigation of model performance and additionally could show the retrieval of challenging driving scenarios within a diverse data foundation.

These capabilities make the proposed metric easily adaptable to different use cases and a flexible tool for a variety of tasks. Additionally, the fine-grained assessment of performance enables a better understanding of specific models' shortcomings thereby bridging the gap towards use in safety-critical applications.

\subsection{Limitations}
The current dataset analysis follows a very simplistic classifying, grouping, and weighting approach without utilizing fine-grained data composition and distribution insights. Additionally, we assume that SOTA models have been only trained on the data stated in the respective technical reports. Especially framework approaches like PatchFusion and DepthAnything, integrating existing pre-trained models, could incorporate more datasets. Similarly, the use of pre-trained backbones could incorporate training data not respected in the evaluations.

The limitations of our metric currently lie within the need for class labels, as well as not integrating class distribution compensation for the comparison between multiple datasets. Sky presents another common limitation in the depth estimation task. Here two ways are common, using the maximum possible distance or zero distance for sky class. However, directly comparing these yields different challenges as model outputs have to be masked.

\subsection{Future Work}
The mentioned limitations offer avenues for future work, also through the advent of object detection foundation models. Here, segmentation models such as SAM 2 \cite{ravi2024sam} could be integrated into a framework-like approach to automatically generate missing class labels. Another aspect is the integration of under- and overestimation distance weighting. One could argue that in the automotive context, distance overestimation is more critical than underestimation. Furthermore, for the evaluation over multiple datasets, class distributions shall be incorporated to reach a unified metric value.

\subsubsection{Acknowledgments} We thank Daniel Bin Schmid of the Technical University of Munich for his valuable insights.


%
%

\newpage
\bibliographystyle{template_files/splncs04}
\bibliography{main}

\begin{thebibliography}{10}
\providecommand{\url}[1]{\texttt{#1}}
\providecommand{\urlprefix}{URL }
\providecommand{\doi}[1]{https://doi.org/#1}

\bibitem{depth_transfer_learning}
Alhashim, I., Wonka, P.: High quality monocular depth estimation via transfer learning. arXiv preprint arXiv:1812.11941  (2018)

\bibitem{dataset_Mapillary_PSD}
Antequera, M.L., Gargallo, P., Hofinger, M., Bulo, S.R., Kuang, Y., Kontschieder, P.: Mapillary planet-scale depth dataset. In: The European Conference Computer Vision (ECCV). pp. 589--604. Springer International Publishing (2020)

\bibitem{beit}
Bao, H., Dong, L., Piao, S., Wei, F.: Beit: Bert pre-training of image transformers. arXiv preprint arXiv:2106.08254  (2021)

\bibitem{dataset_UASOL}
Bauer, Z., Gomez-Donoso, F., Cruz, E., Orts-Escolano, S., Cazorla, M.: Uasol, a large-scale high-resolution outdoor stereo dataset. Scientific data  \textbf{6}(1),  1--14 (2019)

\bibitem{adabins}
Bhat, S.F., Alhashim, I., Wonka, P.: Adabins: Depth estimation using adaptive bins. In: Proceedings of the IEEE/CVF conference on computer vision and pattern recognition. pp. 4009--4018 (2021)

\bibitem{zoedepth}
Bhat, S.F., Birkl, R., Wofk, D., Wonka, P., Müller, M.: Zoedepth: Zero-shot transfer by combining relative and metric depth (2023). \doi{10.48550/ARXIV.2302.12288}, \url{https://arxiv.org/abs/2302.12288}

\bibitem{dataset_Virtual_KITTI_2}
Cabon, Y., Murray, N., Humenberger, M.: Virtual kitti 2. arXiv:2001.10773  (2020)

\bibitem{dataset_nuscenes}
Caesar, H., Bankiti, V., Lang, A.H., Vora, S., Liong, V.E., Xu, Q., Krishnan, A., Pan, Y., Baldan, G., Beijbom, O.: nuscenes: A multimodal dataset for autonomous driving. In: Proc. IEEE Conf. Comp. Vis. Patt. Recogn. pp. 11621--11631 (2020)

\bibitem{dino_v1}
Caron, M., Touvron, H., Misra, I., J{\'e}gou, H., Mairal, J., Bojanowski, P., Joulin, A.: Emerging properties in self-supervised vision transformers. In: Proceedings of the IEEE/CVF international conference on computer vision. pp. 9650--9660 (2021)

\bibitem{dataset_Matterport3d}
Chang, A., Dai, A., Funkhouser, T., Halber, M., Nießner, M., Savva, M., Song, S., Zeng, A., Zhang, Y.: Matterport3d: Learning from rgb-d data in indoor environments. arXiv preprint arXiv:1709.06158  (2017)

\bibitem{loss_edge}
Chen, C., Chen, X., Cheng, H.: On the over-smoothing problem of cnn based disparity estimation. In: Proceedings of the IEEE/CVF International Conference on Computer Vision (ICCV). pp. 8997--9005 (2019)

\bibitem{dataset_DIML}
Cho, J., Min, D., Kim, Y., Sohn, K.: Diml/cvl rgb-d dataset: 2m rgb-d images of natural indoor and outdoor scenes. arXiv: Comp. Res. Repository  (2021)

\bibitem{dataset_Cityscapes}
Cordts, M., Omran, M., Ramos, S., Rehfeld, T., Enzweiler, M., Benenson, R., Franke, U., Roth, S., Schiele, B.: The cityscapes dataset for semantic urban scene understanding. In: Proceedings of the IEEE Conference on Computer Vision and Pattern Recognition (CVPR) (2016)

\bibitem{dataset_ScanNet}
Dai, A., Chang, A.X., Savva, M., Halber, M., Funkhouser, T., Nießner, M.: Scannet: Richlyannotated 3d reconstructions of indoor scenes. In: Proceedings of the IEEE/CVF Conference on Computer Vision and Pattern Recognition (CVPR) (2017)

\bibitem{eigen2015predicting}
Eigen, D., Fergus, R.: Predicting depth, surface normals and semantic labels with a common multi-scale convolutional architecture. In: Proceedings of the IEEE international conference on computer vision. pp. 2650--2658 (2015)

\bibitem{eigen2014depth}
Eigen, D., Puhrsch, C., Fergus, R.: Depth map prediction from a single image using a multi-scale deep network. Advances in neural information processing systems  \textbf{27} (2014)

\bibitem{eisemann_expanding_2020}
Eisemann, L., Froehlich, J., Hartz, A., Maucher, J.: Expanding dynamic range in a single-shot image through a sparse grid of low exposure pixels. Electronic Imaging  \textbf{32}(7) (2020)

\bibitem{eisemann2024opendrive}
Eisemann, L., Maucher, J.: Divide and conquer: A systematic approach for industrial scale high-definition opendrive generation from sparse point clouds. In: 2024 IEEE Intelligent Vehicles Symposium (IV). pp. 2443--2450. IEEE (2024)

\bibitem{GIDAS}
{Federal Highway Research Institute (BASt) and Research Association for Automotive Technology (FAT)}: {GIDAS: German In-Depth Accident Study} (1999), \url{https://www.gidas.org/start-en.html}, accident database collected since 1999

\bibitem{dataset_Virtual_KITTI}
Gaidon, A., Wang, Q., Cabon, Y., Vig, E.: Virtual worlds as proxy for multi-object tracking analysis. In: IEEE Conference on Computer Vision and Pattern Recognition (CVPR) (2016)

\bibitem{dataset_DSEC}
Gehrig, M., Aarents, W., Gehrig, D., Scaramuzza, D.: Dsec: A stereo event camera dataset for driving scenarios. IEEE Robotics and Automation Letters  (2021)

\bibitem{dataset_KITTI}
Geiger, A., Lenz, P., Stiller, C., Urtasun, R.: Vision meets robotics: The kitti dataset. Int. J. Robot. Res.  (2013)

\bibitem{dataset_A2D2}
Geyer, J., Kassahun, Y., Mahmudi, M., Ricou, X., Durgesh, R., Chung, A.S., Hauswald, L., Pham, V.H., M{\"u}hlegg, M., Dorn, S., Fernandez, T., J{\"a}nicke, M., Mirashi, S., Savani, C., Sturm, M., Vorobiov, O., Oelker, M., Garreis, S., Schuberth, P.: {A2D2: Audi Autonomous Driving Dataset}  (2020), \url{https://www.a2d2.audi}

\bibitem{monodepth}
Godard, C., Mac~Aodha, O., Brostow, G.J.: Unsupervised monocular depth estimation with left-right consistency. In: Proceedings of the IEEE conference on computer vision and pattern recognition. pp. 270--279 (2017)

\bibitem{dataset_DDAD}
Guizilini, V., Ambrus, R., Pillai, S., Raventos, A., Gaidon, A.: 3d packing for self-supervised monocular depth estimation. In: Proc. IEEE Conf. Comp. Vis. Patt. Recogn. (2020)

\bibitem{metric3Dv2}
Hu, M., Yin, W., Zhang, C., Cai, Z., Long, X., Chen, H., Wang, K., Yu, G., Shen, C., Shen, S.: Metric3d v2: A versatile monocular geometric foundation model for zero-shot metric depth and surface normal estimation. arXiv preprint arXiv:2404.15506  (2024)

\bibitem{dataset_MVS-Synth}
Huang, P.H., Matzen, K., Kopf, J., Ahuja, N., Huang, J.B.: Deepmvs: Learning multi-view stereopsis. In: CVPR (2018)

\bibitem{marigold}
Ke, B., Obukhov, A., Huang, S., Metzger, N., Daudt, R.C., Schindler, K.: Repurposing diffusion-based image generators for monocular depth estimation. In: Proceedings of the IEEE/CVF Conference on Computer Vision and Pattern Recognition (CVPR) (2024)

\bibitem{dataset_Lyft}
Kesten, R., Usman, M., Houston, J., Pandya, T., Nadhamuni, K., Ferreira, A., Yuan, M., Low, B., Jain, A., Ondruska, P., Omari, S., Shah, S., Kulkarni, A., Kazakova, A., Tao, C., Platinsky, L., Jiang, W., Shet, V.: Level 5 perception dataset 2020  (2019), https://level-5.global/level5/data/

\bibitem{dataset_DIML-Indoor}
Kim, Y., Jung, H., Min, D., Sohn, K.: Deep monocular depth estimation via integration of global and local predictions. IEEE transactions on Image Processing  \textbf{27}(8),  4131--4144 (2018)

\bibitem{alexnet}
Krizhevsky, A., Sutskever, I., Hinton, G.E.: Imagenet classification with deep convolutional neural networks. In: Advances in Neural Information Processing Systems. vol.~25. Curran Associates, Inc. (2012), \url{https://proceedings.neurips.cc/paper_files/paper/2012/file/c399862d3b9d6b76c8436e924a68c45b-Paper.pdf}

\bibitem{dataset_MegaDepth}
Li, Z., Snavely, N.: Megadepth: Learning singleview depth prediction from internet photos. In: CVPR (2018)

\bibitem{patchfusion}
Li, Z., Bhat, S.F., Wonka, P.: Patchfusion: An end-to-end tile-based framework for high-resolution monocular metric depth estimation  (2024)

\bibitem{dataset_goose}
Mortimer, P., Hagmanns, R., Granero, M., Luettel, T., Petereit, J., Wuensche, H.J.: The goose dataset for perception in unstructured environments  (2024), \url{https://arxiv.org/abs/2310.16788}

\bibitem{dino_v2}
Oquab, M., Darcet, T., Moutakanni, T., Vo, H., Szafraniec, M., Khalidov, V., Fernandez, P., Haziza, D., Massa, F., El-Nouby, A., et~al.: Dinov2: Learning robust visual features without supervision. arXiv preprint arXiv:2304.07193  (2023)

\bibitem{metric_point_clouds}
Ornek, E.P., Mudgal, S., Wald, J., Wang, Y., Navab, N., Tombari, F.: From 2d to 3d: Rethinking benchmarking of monocular depth prediction. arXiv preprint arXiv:2203.08122  (2022)

\bibitem{ecodepth}
Patni, S., Agarwal, A., Arora, C.: Ecodepth: Effective conditioning of diffusion models for monocular depth estimation. In: Proceedings of the IEEE/CVF Conference on Computer Vision and Pattern Recognition (CVPR). pp. 28285--28295 (June 2024)

\bibitem{unidepth}
Piccinelli, L., Yang, Y.H., Sakaridis, C., Segu, M., Li, S., Van~Gool, L., Yu, F.: {U}ni{D}epth: Universal monocular metric depth estimation. In: Proceedings of the IEEE/CVF Conference on Computer Vision and Pattern Recognition (CVPR) (2024)

\bibitem{ros_stanford}
Quigley, M.: Ros: an open-source robot operating system. In: IEEE International Conference on Robotics and Automation (2009), \url{https://api.semanticscholar.org/CorpusID:6324125}

\bibitem{dataset_HM3d}
Ramakrishnan, S.K., Gokaslan, A., Wijmans, E., Maksymets, O., Clegg, A., Turner, J., Undersander, E., Galuba, W., Westbury, A., A.~X.~Chang, e.a.: Habitat-matterport 3d dataset (hm3d): 1000 large-scale 3d environments for embodied ai. arXiv preprint arXiv:2109.08238  (2021)

\bibitem{dataset_3D_Movies}
Ranftl, R., Lasinger, K., Hafner, D., Schindler, K., Koltun, V.: Towards robust monocular depth estimation: Mixing datasets for zero-shot cross-dataset transfer. IEEE Transactions on Pattern Analysis and Machine Intelligence (TPAMI)  (2020)

\bibitem{midas}
Ranftl, R., Lasinger, K., Hafner, D., Schindler, K., Koltun, V.: Towards robust monocular depth estimation: Mixing datasets for zero-shot cross-dataset transfer. IEEE Transactions on Pattern Analysis and Machine Intelligence  \textbf{44}(3) (2022)

\bibitem{ravi2024sam}
Ravi, N., Gabeur, V., Hu, Y.T., Hu, R., Ryali, C., Ma, T., Khedr, H., R{\"a}dle, R., Rolland, C., Gustafson, L., et~al.: Sam 2: Segment anything in images and videos. arXiv preprint arXiv:2408.00714  (2024)

\bibitem{dataset_Hypersim}
Roberts, M., Ramapuram, J., Ranjan, A., Kumar, A., Bautista, M.A., Paczan, N., Webb, R., Susskind, J.M.: Hypersim: A photorealistic synthetic dataset for holistic indoor scene understanding. pp. 10912--10922 (2021)

\bibitem{conditioned_diff_model}
Saxena, S., Hur, J., Herrmann, C., Sun, D., Fleet, D.J.: Zero-shot metric depth with a field-of-view conditioned diffusion model (2023)

\bibitem{dataset_middlebury}
Scharstein, D., Hirschm{\"u}ller, H., Kitajima, Y., Krathwohl, G., Ne{\v{s}}i{\'c}, N., Wang, X., Westling, P.: High-resolution stereo datasets with subpixel-accurate ground truth. In: Pattern Recognition: 36th German Conference, GCPR 2014, M{\"u}nster, Germany, September 2-5, 2014, Proceedings 36. pp. 31--42. Springer (2014)

\bibitem{dataset_NYU_Depth_V2}
Silberman, N., Hoiem, D., Kohli, P., Fergus, R.: Indoor segmentation and support inference from rgbd images. In: Proc. Eur. Conf. Comp. Vis., pp. 746--760. Springer (2012)

\bibitem{loss_median_absolute_deviation_normalization}
Singh, D., Singh, B.: Investigating the impact of data normalization on classification performance. Applied Soft Computing  (2019)

\bibitem{dataset_Replica}
Straub, J., Whelan, T., Ma, L., Chen, Y., Wijmans, E., Green, S., Engel, J.J., Mur-Artal, R., Ren, C., Verma, S., et~al.: The replica dataset: A digital replica of indoor spaces. arXiv preprint arXiv:1906.05797  (2019)

\bibitem{dataset_Waymo}
Sun, P., Kretzschmar, H., Dotiwalla, X., Chouard, A., Patnaik, V., Tsui, P., Guo, J., Zhou, Y., Chai, Y., Benjamin~Caine, e.a.: Scalability in perception for autonomous driving: Waymo open dataset. In: Proceedings of the IEEE/CVF Conference on Computer Vision and Pattern Recognition (CVPR). pp. 2446--2454 (2020)

\bibitem{FE_BorderFollowing}
Suzuki, S., et~al.: Topological structural analysis of digitized binary images by border following. Computer vision, graphics, and image processing  \textbf{30}(1),  32--46 (1985)

\bibitem{loss_edge2}
Tosi, F., Liao, Y., Schmitt, C., Geiger, A.: Smd-nets: Stereo mixture density networks. In: Proceedings of the IEEE/CVF Conference on Computer Vision and Pattern Recognition (CVPR). pp. 8942--8952 (2021)

\bibitem{dataset_UnrealStereo4K}
Tosi, F., Liao, Y., Schmitt, C., Geiger, A.: Smd-nets: Stereo mixture density networks. In: CVPR. pp. 8942--8952 (2021)

\bibitem{dataset_kitti_benchmark}
Uhrig, J., Schneider, N., Schneider, L., Franke, U., Brox, T., Geiger, A.: Sparsity invariant cnns. In: International Conference on 3D Vision (3DV) (2017)

\bibitem{dataset_DIODE}
Vasiljevic, I., Kolkin, N., Zhang, S., Luo, R., Wang, H., Dai, F.Z., Daniele, A.F., Mostajabi, M., Basart, S., Walter, M.R., Shakhnarovich, G.: Diode: A dense indoor and outdoor depth dataset. CoRR  \textbf{abs/1908.00463} (2019)

\bibitem{transformer}
Vaswani, A., Shazeer, N., Parmar, N., Uszkoreit, J., Jones, L., Gomez, A.N., Kaiser, {\L}., Polosukhin, I.: Attention is all you need. Advances in neural information processing systems  \textbf{30} (2017)

\bibitem{dataset_WSVD}
Wang, C., Lucey, S., Perazzi, F., Wang, O.: Web stereo video supervision for depth prediction from dynamic scenes. In: 2019 International Conference on 3D Vision (3DV). pp. 348--357. IEEE (2019)

\bibitem{dataset_ApolloScape}
Wang, P., Huang, X., Cheng, X., Zhou, D., Geng, Q., Yang, R.: The apolloscape open dataset for autonomous driving and its application. IEEE Transactions on Pattern Analysis and Machine Intelligence  \textbf{PP}, ~1--1 (07 2019). \doi{10.1109/TPAMI.2019.2926463}

\bibitem{dataset_IRS}
Wang, Q., Zheng, S., Yan, Q., Deng, F., Zhao, K., Chu, X.: Irs: A large naturalistic indoor robotics stereo dataset to train deep models for disparity and surface normal estimation. In: ICME (2021)

\bibitem{dataset_TartanAir}
Wang, W., Zhu, D., Wang, X., Hu, Y., Qiu, Y., Wang, C., Hu, Y., Kapoor, A., Scherer, S.: Tartanair: A dataset to push the limits of visual slam. In: IROS (2020)

\bibitem{dataset_Argoverse2}
Wilson, B., Qi, W., Agarwal, T., Lambert, J., Singh, J., Khandelwal, S., Pan, B., Kumar, R., Hartnett, A., Pontes, J.K., Ramanan, D., Carr, P., Hays, J.: Argoverse 2: Next generation datasets for self-driving perception and forecasting. In: Advances in Neural Information Processing Systems (2021)

\bibitem{dataset_ReDWeb}
Xian, K., Shen, C., Cao, Z., Lu, H., Xiao, Y., Li, R., Luo, Z.: Monocular relative depth perception with web stereo data supervision. In: Proceedings of the IEEE Conference on Computer Vision and Pattern Recognition. pp. 311--320 (2018)

\bibitem{dataset_HRWSI}
Xian, K., Zhang, J., Wang, O., Mai, L., Lin, Z., Cao, Z.: Structure-guided ranking loss for single image depth prediction. In: CVPR (2020)

\bibitem{dataset_Pandaset}
Xiao, P., Shao, Z., Hao, S., Zhang, Z., Chai, X., Jiao, J., Li, Z., Wu, J., Sun, K., Jiang, K., Wang, Y., Yang, D.: Pandaset: Advanced sensor suite dataset for autonomous driving. In: IEEE Int. Intelligent Transportation Systems Conf. (2021)

\bibitem{dataset_Driving_Stereo}
Yang, G., Song, X., Huang, C., Deng, Z., Shi, J., Zhou, B.: Drivingstereo: A large-scale dataset for stereo matching in autonomous driving scenarios. In: Proceedings of the IEEE/CVF Conference on Computer Vision and Pattern Recognition (CVPR) (2019)

\bibitem{depthanything}
Yang, L., Kang, B., Huang, Z., Xu, X., Feng, J., Zhao, H.: Depth anything: Unleashing the power of large-scale unlabeled data. In: CVPR (2024)

\bibitem{depth_anything_v2}
Yang, L., Kang, B., Huang, Z., Zhao, Z., Xu, X., Feng, J., Zhao, H.: Depth anything v2. arXiv:2406.09414  (2024)

\bibitem{dataset_BlendedMVS}
Yao, Y., Luo, Z., Li, S., Zhang, J., Ren, Y., Zhou, L., Fang, T., Quan, L.: Blendedmvs: A largescale dataset for generalized multi-view stereo networks. In: CVPR (2020)

\bibitem{metric3Dv1}
Yin, W., Zhang, C., Chen, H., Cai, Z., Yu, G., Wang, K., Chen, X., Shen, C.: Metric3d: Towards zero-shot metric 3d prediction from a single image (2023)

\bibitem{dataset_BDD100K}
Yu, F., Chen, H., Wang, X., Xian, W., Chen, Y., Liu, F., Madhavan, V., Darrell, T.: Bdd100k: A diverse driving dataset for heterogeneous multitask learning. In: Proceedings of the IEEE/CVF Conference on Computer Vision and Pattern Recognition (CVPR). pp. 2636--2645 (2020)

\bibitem{dataset_Taskonomy}
Zamir, A.R., Sax, A., Shen, W.B., Guibas, L., Malik, J., Savarese, S.: Taskonomy: Disentangling task transfer learning. In: Proceedings of the IEEE/CVF Conference on Computer Vision and Pattern Recognition (CVPR). IEEE (2018)

\end{thebibliography}
\end{document}